# Evaluating the Impact of Lab Test Results on Large Language Models Generated Differential Diagnoses from Clinical Case Vignettes

Balu Bhasuran, PhD[1], Qiao Jin, MD[2], Yuzhang Xie, MS[3], Carl Yang, PhD[3], Karim Hanna, MD[4], Jennifer Costa, MD[4], Cindy Shavor, MD[4], Zhiyong Lu, PhD[2], Zhe He[1, *], PhD[1]

[1]Florida State University, Tallahassee, FL; [2]National Center for Biotechnology Information, National Library of Medicine, Bethesda, MD; [3]Emory University, Atlanta, GA; [4]Morsani College of Medicine, University of South Florida, Tampa, FL

**\*Corresponding author:**

Zhe He, PhD, FAMIA

School of Information

Florida State University

142 Collegiate Loop

Tallahassee, Florida, 32306-2100

United States

Email: zhe@fsu.edu

Phone: (001) 850-644-5775

**Abstract**

Differential diagnosis is crucial for medicine as it helps healthcare providers systematically distinguish between conditions that share similar symptoms. This study assesses the impact of lab test results on differential diagnoses (DDx) made by large language models (LLMs). Clinical vignettes from 50 case reports from PubMed Central were created incorporating patient demographics, symptoms, and lab results. Five LLMs—GPT-4, GPT-3.5, Llama-2-70b, Claude-2, and Mixtral-8x7B were tested to generate Top 10, Top 5, and Top 1 DDx with and without lab data. A comprehensive evaluation involving GPT-4, a knowledge graph, and clinicians was conducted. GPT-4 performed best, achieving 55% accuracy for Top 1 diagnoses and 60% for Top 10 with lab data, with lenient accuracy up to 80%. Lab results significantly improved accuracy, with GPT-4 and Mixtral excelling, though exact match rates were low. Lab tests, including liver function, metabolic/toxicology panels, and serology/immune tests, were generally interpreted correctly by LLMs for differential diagnosis.

# Introduction

Accurate diagnosis is critical for the effective management of patients' conditions, as it directly influences treatment decisions and overall patient outcomes[1]. A correct diagnosis ensures that patients receive timely and appropriate interventions, which not only improves outcomes but also reduces morbidity and mortality. Moreover, a correct diagnosis enables healthcare providers to select the most effective therapies, minimizing the risks associated with unnecessary or inappropriate treatments. By reducing diagnostic errors, accurate diagnosis streamlines patient care, eliminating the need for excessive or repeated testing, and ultimately lowering healthcare costs through reduced hospital stays and unnecessary procedures. Furthermore, patient safety is enhanced when accurate diagnoses mitigate the risks of complications and adverse drug reactions.[2,3].

In contrast, differential diagnosis (DDx) plays a crucial role in enhancing clinical decision-making by systematically evaluating and ruling in/out potential conditions. The prediction of differential diagnoses provides several advantages, such as improving the reliability of clinical data, supporting the discovery of new treatments, and fostering a better understanding of diseases. Differential diagnosis also serves as an essential learning tool for medical students and healthcare professionals, enhancing their diagnostic skills. Its application spans a wide range of diseases including Alzheimer's disease[4], multiple sclerosis[5], inflammatory bowel disease colitis[6], epilepsy[7], stroke[8], and others.

Clinicians often rely on their expertise and various case presentations to achieve diagnostic excellence. Differential diagnosis has long been a necessary step in clinical settings, prompting the development of earlier systems such as differential diagnosis generators[9] and symptom checkers[10]. High-performing deep learning models have also been created for generating DDx in various specialties, including radiology[11], ophthalmology[12] and dermatology[13]. However, these systems face significant issues: they often require structured data, lack the ability to provide valid reasoning for differential diagnoses, and do not have any interactive capabilities. The recent emergence of large language models (LLMs) such as OpenAI's Generative Pretrained Transformers (GPT-4)[14] offers significant potential for developing tools that aid in generating accurate DDx. Previous studies have shown that LLMs can generate DDx with satisfactory performance.

Kanjee et al. conducted one of the earliest studies evaluating LLMs for predicting differential diagnoses from clinical cases[15]. They tested GPT-4 on 70 cases from the New England Journal of Medicine (NEJM) Clinicopathological Conference (CPC), using a differential quality score ranging from 0 (no diagnosis) to 5 (exact diagnosis). The study found that GPT-4 achieved a final diagnosis in 39% (27/70) of cases and included the correct diagnosis in the differential list in 64% (45/70) cases. McDuff et al. introduced an optimized LLM for differential diagnosis evaluation using 302 clinical cases from the same NEJM CPC[16]. Their PaLM-2-based LLM demonstrated superior standalone performance compared to unassisted clinicians, with a top-10 accuracy of 59.1% versus 33.6%. In this benchmark, they also outperformed GPT-4 in both top-1 and top-10 accuracy.

Following the success of LLM-based DDx from general case reports corpus, researchers applied this approach to various specialized domains, such as rheumatology[17], neurodegenerative disorders[18], autoinflammatory disorders[19], and pediatric critical care[20]. Krusche et al. compared the performance of GPT-4 in diagnosing rheumatologic conditions to that of rheumatologists, reporting comparable accuracy with the correct diagnosis as the top diagnosis in 35% versus 39% of cases, and among the top 3 diagnoses in 60% versus 55% of cases[17]. Koga et al. evaluated ChatGPT-3.5, ChatGPT-4, and Google Bard (now called Gemini) in predicting neuropathologic diagnoses from 25 clinical summaries[18]. The models correctly made primary diagnoses in 32%, 52%, and 40% of cases, respectively, and included the correct diagnoses among the differential diagnosis in 76%, 84%, and 76% of cases, respectively. Pillai and Pillai assessed the diagnostic accuracy of GPT-3.5, GPT-4, and LLaMA for autoinflammatory disorders, specifically focusing on Deficiency of Interleukin-1 Receptor Antagonist (DIRA) and Familial Mediterranean Fever (FMF), using 40 clinical vignettes[19]. They reported that all three models had higher accuracy in identifying FMF compared to DIRA, with GPT-4 correctly identifying 65% of FMF patients versus 90% by clinicians, and 30% of DIRA patients versus 60% by clinicians. Interestingly, they reported that LLaMA 2 had 0% accuracy in identifying DIRA patients. Apart from using a small number of clinical cases, Akhondi et al. used two general language models (BioGPT-Large and LLaMa-65B) and two fine-tuned models (fine-tuned BioGPT-Large and fine-tuned LLaMa-7B) for differential diagnosis in a pediatric critical care setting[20]. The models were generated using 1,916,538 clinical notes from 32,454 unique patients, evaluated the models using mixed methods regression finding that the differential diagnoses generated by clinicians and the fine-tuned LLaMa-7B were ranked highest in quality in 144 (55%) and 74 (29%) cases, respectively. Hirosawa et al. evaluated the accuracy of differential diagnosis lists generated by GPT-3.5 and GPT-4 using 53 case reports from a general internal medicine (GIM) department. The study found that GPT-4 achieved 83% accuracy in the top 10 predictions, 81% in the top 5, and 60% in the top 1 diagnosis. GPT-4's performance was comparable to that of physicians, with accuracy rates of 83% vs. 75% for the top 10, 81% vs. 67% for the top 5, and 60% vs. 50% for the top diagnosis. Additionally, the study reported no significant difference in diagnosis accuracy based on the open access status or the publication date (before 2011 vs. 2022)[21].

While existing studies evaluate the accuracy of differential diagnosis derived from case reports by LLMs, the role of lab test results in these predictions remains unknown. As lab results play a crucial role in diagnosis, we hypothesize that lab test results enhance the accuracy of differential diagnoses by supplying critical information about a patient's physiological status, which may not be evident from symptoms alone. This study aims to evaluate the role of lab test results in improving the accuracy of differential diagnosis when applying five large language models to clinical case reports collected from the PMC-Patients dataset[22], a publicly available benchmark dataset that contains patient summaries and relationships extracted from PubMed Central (PMC) articles. This novel dataset is collected from case reports in PMC along with the PubMed citation graph.

Our key contributions are summarized as follows:

1. To the best of our knowledge, we are reporting for the first time the role of lab test results in improving the accuracy of differential diagnosis predictions using LLMs.
2. Evaluating the performance of both proprietary and open-source LLMs on 50 challenging diagnostic cases from published case report studies.

3. Reporting that a combination of Biomedical Knowledge Graphs and GPT-4 (BKG-GPT) can perform automatic assessments of DDx with a level of accuracy comparable to that of clinicians.
4. Conducting comprehensive error analysis to reveal a deeper understanding of LLMs' DDx predictions.

# Results

## Method Overview

We evaluated the impact of laboratory test results on improving the accuracy of differential diagnosis using five large language models: GPT -4[14], GPT -3.5[23], Llama-2[24], Claude2[25], and Mixtral[26]. Clinical case reports for this assessment were obtained from the PMC-Patients dataset. From 50 selected case reports, we manually generated clinical vignettes that included details such as patient age, gender, symptoms, laboratory test results, and other relevant information, allowing the models to generate differential diagnosis responses. The clinical vignettes used for this study is described in detail in the Methods section (see Methods: Clinical Vignettes). A specific prompt was designed to instruct the models to consider all relevant details and provide differential diagnoses, including Top 1, Top 5, and Top 10 DDx lists. Model predictions were reviewed by clinicians and automatically evaluated using a knowledge graph and GPT-4, utilizing exact match, relevance, and incorrect predictions. The diagnostic accuracy was evaluated using accuracy and lenient accuracy metrics for Top 1, Top 5, and Top 10 DDx, derived from clinical vignettes with and without laboratory test data. Accuracy is calculated by assigning weights of 1.0 for exact matches, 0.5 (or 0.75 for lenient accuracy) for relevant matches, and 0.0 for incorrect matches, with the sum divided by the total of 50 diagnoses evaluated. For the evaluation metrics used, see Methods: Evaluation of Differential Diagnosis Lists. An overview of the study pipeline is presented in Fig. 1.

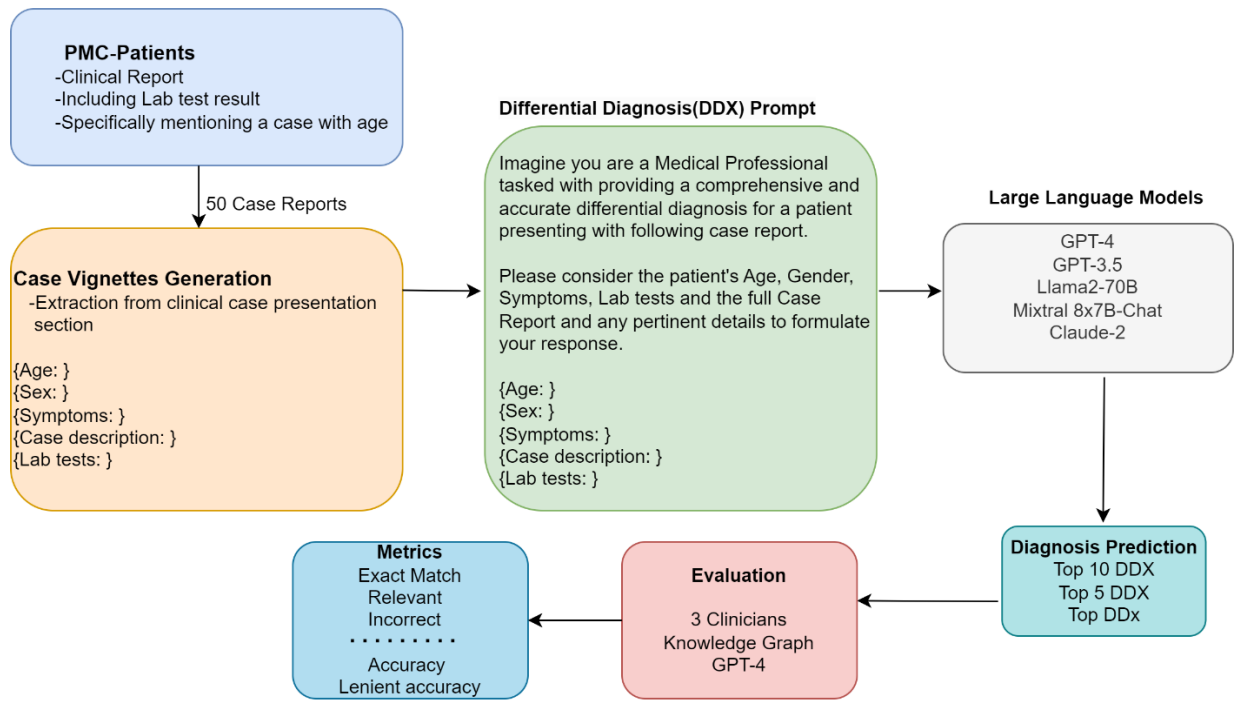

**Fig. 1. Schematic architecture of the study pipeline**

The study leveraged 50 clinical case reports spanning clinical categories of Endocrine/Metabolic, Cardiovascular, Hematologic/Oncologic, Infectious Diseases, Neurological Disorders, Gastrointestinal and Hepatic Conditions, Renal Conditions, Urological Conditions, Toxicological Conditions, Musculoskeletal Disorders, Autoimmune Disorders and Hematologic/Coagulation Disorders. Fig. 2 illustrates the distribution of diseases across various clinical categories. The case reports included 29 male and 21 female patients with the median age at approximately 45. The ages in the reports range from a minimum of 1 to a maximum of 79.

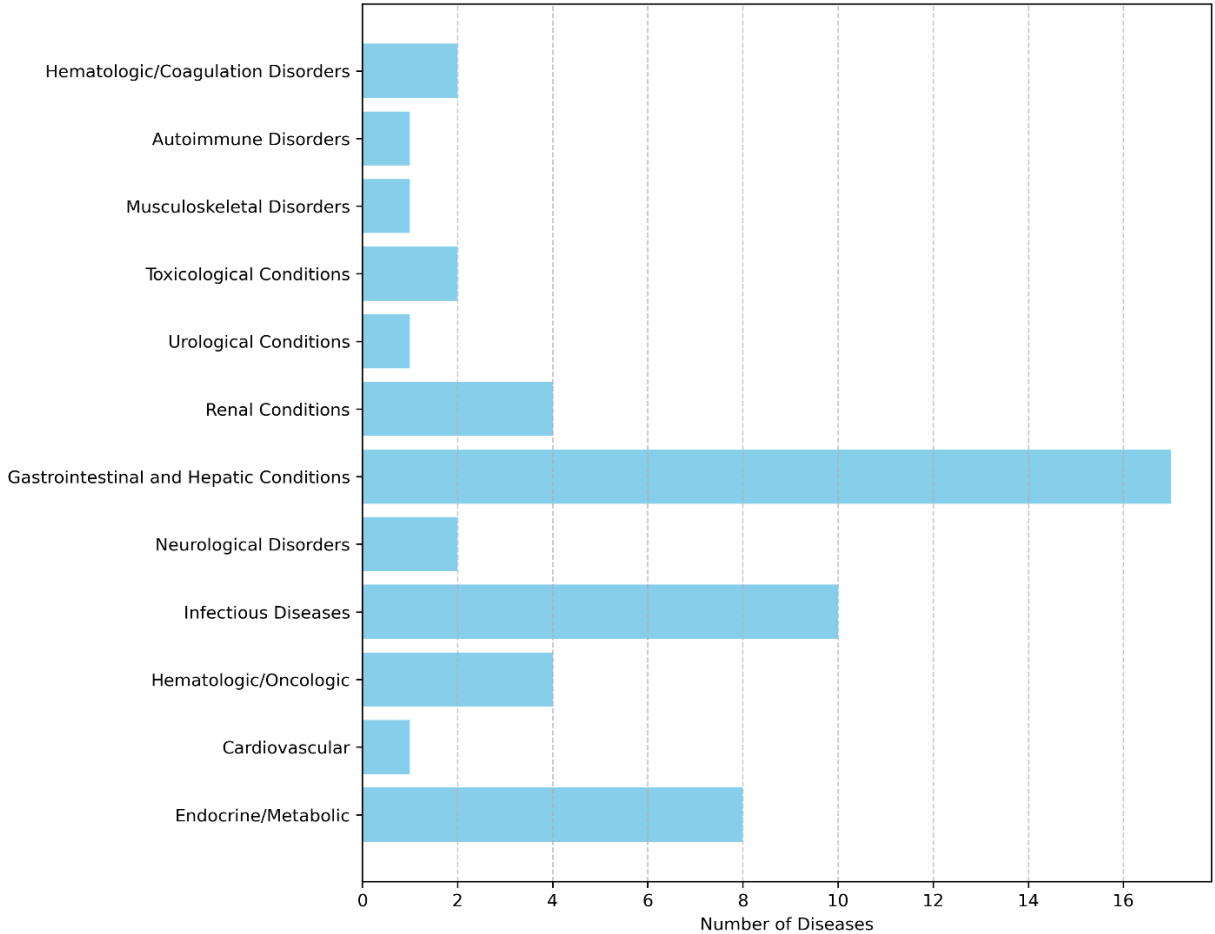

**Fig. 2. Clinical categories of 50 case reports across various medical conditions.**

Table 1 presents an overview of studies employing LLMs for DDx across various medical specialties. These studies, sourced from reputable publications, utilize models such as GPT-4, GPT-3.5, PaLM-2, Llama-2, and BioGPT-Large. Evaluation in these studies is primarily conducted by expert clinicians or specialists, ensuring diagnostic accuracy and clinical relevance. The current study advances this field by leveraging a diverse PMC patient dataset to address a broad range of medical conditions and integrates models like GPT-4, GPT-3.5, Llama-2, Claude2, and Mixtral. A key technical contribution of this work is the introduction of evaluation using a biomedical knowledge graph (BKG). With GPT-4 and BKG integration (BKG-GPT), the system can perform automatic assessments with a level of accuracy comparable to that of clinicians, enhancing both interpretability and scalability. This automated capability represents a significant step toward streamlining clinical decision-making.

**Table 1: Different LLM based studies for differential diagnosis**

| Study | Case Report Size, n | Source | Specialty | LLMs | Evaluation |
|---|---|---|---|---|---|
| Kanjee et al., 2023[15] | 70 | New England Journal of Medicine clinicopathological case conferences | Pathology | GPT-4 | Clinician |
| McDuff, D. et al, 2023[16] | 302 | New England Journal of Medicine clinicopathological case conferences | Pathology | PaLM-2 Large, GPT-4 | Clinician, GPT-4 |
| Akhondi et al., 2023[20] | 132 notes | Pediatric Intensive Care Unit (PICU) admission notes | Pediatric Intensive Care | BioGPT-Large, Llama-65B, Llama-7B | Clinician |
| Krusche et al., 2023[17] | 20 | Evaluation of Triage Tools in Rheumatology (bETTeR) | Rheumatology | GPT-4 | Rheumatologist |
| Koga et al., 2023[18] | 25 | Mayo Clinic brain bank Clinico-Pathological Conferences | Neuropathology | GPT-3.5, GPT-4, and Google Bard | Neuropathologist |
| Pillai et al., 2023[19] | 40 | PubMed database | Autoinflammatory disorders | GPT-3.5, GPT-4, and Llama-2 | Internal medicine physician |
| Hirosawa et al., 2023[21] | 52 | PubMed database | General Internal Medicine (GIM) | GPT-3.5, GPT-4 | Internal medicine physician |
| Current Study | 50 | PMC-Patients dataset | Gastrointestinal and Hepatic Conditions, Renal Conditions, Urological Conditions, Toxicological Conditions, Musculoskeletal Disorders, Autoimmune Disorders etc. | GPT-4, GPT-3.5, Llama-2, Claude2, and Mixtral | Clinician, GPT-4, Knowledge Graph |

**Diagnostic Performance**

The current study generated 1,500 DDx sets from 50 case reports, each assessed by five LLMs across six conditions (Top 1, Top 5 and Top 10, with and without lab test results). To evaluate the accuracy of these LLM-generated DDx sets, a two-stage evaluation approach was adopted. In Stage 1, 20% of the total DDx list (300 diagnoses predictions) from 10 case reports were selected and assessed by clinicians, followed by a comparison with a knowledge graph, and GPT-4. Stage 2 involved evaluating all 1,500 DDx sets using a combined approach of the knowledge graph and GPT-4. This multi-layered evaluation aimed to provide a comparative assessment of the LLMs' ability to generate accurate differential diagnoses.

**Evaluation of 300 Differential Diagnoses by Clinicians**

The clinician-based evaluation was limited to 300 differential diagnoses (a random 20% of the total, from 10 case reports) due to the significant time and effort required from clinicians, including the detailed assessment of lab test contributions. Ten clinical vignettes based DDx were used for clinician evaluation, with three clinicians reviewing different result sets. The clinicians were asked to predict a provided pair of actual and LLM-predicted diagnoses as exact, relevant, or incorrect. Table 2 provides a detailed performance comparison of all the LLMs. GPT-3.5 and GPT-4 generally performed better, with GPT-3.5 reaching an exact match accuracy of 80% for the top 1 DDx with lab data, and GPT-4 achieving 75% accuracy for the top 10 DDx with lab data. Lenient accuracy was higher across models, with GPT-4 achieving 88% for the top 10 DDx with lab results. Including lab test data improved DDx predictions for all models, highlighting the importance of lab information in enhancing diagnostic accuracy. This manual evaluation was compared with a knowledge graph and GPT-4 to explore the feasibility of a fully automated evaluation, assessed by aligning the predictions of different combinations of the knowledge graph and GPT-4 with the clinician-provided evaluation.

**Table 2.** The accuracy of five LLMs in generating differential diagnosis lists of Top 1, Top 5, and Top 10 from 10 case reports evaluated by clinicians. Accuracies and lenient accuracy were calculated using equations 1 and 2 respectively.

| LLM | Top 1 Differential Diagnosis (with lab) | | | | | Top 1 Differential Diagnosis (without lab) | | | | |
|---|---|---|---|---|---|---|---|---|---|---|
| | Exact Match | Relevant | Incorrect | Accuracy | Lenient accuracy | Exact Match | Relevant | Incorrect | Accuracy | Lenient accuracy |
| Llama-2 | 3 | 6 | 1 | 60% | 75% | 2 | 4 | 4 | 40% | 50% |
| Claude-2 | 5 | 4 | 1 | 70% | 80% | 2 | 4 | 4 | 40% | 50% |
| Mixtral | 3 | 6 | 1 | 60% | 75% | 1 | 6 | 3 | 40% | 55% |
| GPT-3.5 | 7 | 2 | 1 | 80% | 85% | 2 | 6 | 2 | 50% | 65% |
| GPT-4 | 4 | 6 | 0 | 70% | 85% | 2 | 6 | 2 | 50% | 65% |
|  | Top 5 Differential Diagnosis (with lab) | | | | | Top 5 Differential Diagnosis (without lab) | | | | |
| Llama-2 | 2 | 6 | 2 | 50% | 65% | 1 | 7 | 2 | 45% | 63% |
| Claude-2 | 3 | 6 | 1 | 60% | 75% | 1 | 8 | 1 | 50% | 70% |
| Mixtral | 2 | 7 | 1 | 55% | 73% | 1 | 8 | 1 | 50% | 70% |
| GPT-3.5 | 4 | 4 | 2 | 60% | 70% | 1 | 8 | 1 | 50% | 70% |
| GPT-4 | 4 | 6 | 0 | 70% | 85% | 1 | 8 | 1 | 50% | 70% |
|  | Top 10 Differential Diagnosis (with lab) | | | | | Top 10 Differential Diagnosis (without lab) | | | | |
| Llama-2 | 2 | 6 | 2 | 50% | 65% | 1 | 7 | 2 | 45% | 63% |
| Claude-2 | 4 | 5 | 1 | 65% | 78% | 2 | 7 | 1 | 55% | 73% |
| Mixtral | 3 | 6 | 1 | 60% | 75% | 1 | 7 | 2 | 45% | 63% |
| GPT-3.5 | 4 | 5 | 1 | 65% | 78% | 2 | 7 | 1 | 55% | 73% |
| GPT-4 | 5 | 5 | 0 | 75% | 88% | 1 | 8 | 1 | 50% | 70% |

**Stage 1: Evaluation of 300 Differential Diagnoses by Clinicians, Knowledge Graph, and GPT-4**

This evaluation aimed to determine whether automated evaluations by BKG and GPT4 aligned with clinician evaluations and was carried out through four different scenarios: GPT-4 vs. Clinicians, GPT-4 vs. BKG, Clinicians vs. BKG, and Clinicians vs. the combined GPT-4+KG (BKG-GPT). Table 3 provides a detailed comparison of the agreement and disagreement percentages across four different evaluation scenarios involving five LLMs.

In the first scenario (GPT-4 vs. Clinicians), predictions from LLMs were evaluated by comparing their outputs with both GPT-4 and clinician evaluations. The results highlight varying degrees of alignment. Claude-2 achieved an alignment percentage of 75% with GPT-4 and clinicians, showing a relatively high alignment with a variance percentage of 25%. GPT-3.5 had a slightly lower alignment percentage of 71.67%, with a variance of 28.33%. GPT-4, when compared with clinicians, demonstrated a 73.33% alignment percentage and a variance of 26.67%. LLaMa-2 had a lower alignment percentage of 66.67% and a variance of 33.33%, indicating more divergence from clinician evaluations. Mixtral's alignment percentage was 73.33%, identical to

GPT-4, with a variance of 26.67%. The average alignment percentage for GPT-4 and clinicians across these LLMs was 72%, with a variance of 28%.

In the second scenario (GPT-4 vs. BKG), the predictions of the LLMs were compared between GPT-4 and the BKG predictions. The alignment percentages in this context varied. Claude-2 showed a moderate alignment percentage of 65%, with a variance percentage of 35%. GPT-3.5 performed the best in this scenario, achieving the highest alignment percentage of 86.67% and a variance of 13.33%. GPT-4 also showed strong alignment with BKG evaluations, with an alignment percentage of 78.33% and a variance of 21.67%. LLaMa-2 had the lowest alignment percentage at 56.67%, indicating more substantial divergence from BKG evaluations, with a variance percentage of 43.33%. Mixtral had an alignment percentage of 68.33% and a variance of 31.67%. The average alignment percentage for GPT-4 and KG was 71%, with a variance of 29%.

In the third scenario (Clinicians vs. BKG), Claude-2 achieved an alignment percentage of 80%, with a variance percentage of 20%, showing strong alignment between clinician and BKG evaluations. GPT-3.5 had a slightly higher alignment percentage at 81.67%, with a variance of 18.33%. GPT-4 performed the best in this scenario, with the highest alignment percentage of 91.67% and a variance of 8.33%, indicating very close alignment between GPT-4's predictions and BKG evaluations. LLaMa-2 showed a lower alignment percentage of 73.33%, with a variance of 26.67%. Mixtral had a strong performance with an alignment percentage of 85% and a variance percentage of 15%. The average alignment percentage between clinicians and BKG across the LLMs was 82.33%, with a variance of 17.66%. In all scenarios, "agreement" refers to the match between the LLMs' predictions and the evaluations from GPT-4, clinicians, and BKG, while "disagreement" reflects mismatches in prediction output evaluations.

In the final scenario (Clinicians vs BKG-GPT) a combined mean score was then used to classify predictions into one of three categories: Exact Match, Relevant, or Incorrect. An Exact Match was defined as a mean score of 0.75 or 1, indicating high alignment with the correct diagnosis. A Relevant prediction, with a score of 0.25 or 0.5, indicated partial accuracy and some useful information. An Incorrect prediction, scored at 0, indicated a failure to provide an accurate diagnosis by both models. The results demonstrated that this approach yielded the highest accuracies across all models. Claude-2 achieved 83.33% accuracy, while GPT-3.5 reached 86.67%. GPT-4 had the highest accuracy overall at 93.33%, showcasing exceptional alignment with clinician judgments. LLaMa-2-70b had an accuracy of 80%, showing significant alignment but lower than GPT-3.5 and GPT-4. Mixtral-8x7B performed similarly to GPT-3.5 with 86.67% accuracy. Overall, this scenario highlighted strong performance across models, particularly GPT-4, in aligning with clinicians' evaluations. The average match count was 86, with an average mismatch count of 14 across the models.

This analysis reveals that GPT-4, particularly when combined with BKG, consistently shows the highest alignment percentages with clinician evaluations, achieving the strongest performance across most scenarios. GPT-3.5 also performs well, especially in the BKG-related evaluations. LLaMa-2 consistently shows lower alignment percentages, indicating less alignment with both clinician and BKG evaluations. Mixtral generally performs well, particularly in the combined GPT-4 and BKG evaluation context. This detailed assessment underscores the utility of integrating predictions from LLMs like GPT-4 with BKG evaluations to improve alignment in medical decision-making contexts.

**Table 3. Comparative Accuracy and Mismatch Analysis of 300 Predictions Across GPT-4, Biomedical Knowledge Graph, and Clinician Evaluations**

| Evaluation | Agreement | Disagreement | Alignment Percentage (%) | Variance Percentage (%) |
|---|---|---|---|---|
| **GPT-4 vs Clinician (First Scenario)** | | | | |
| Claude | 45 | 15 | **75.00** | 25.00 |
| GPT-3.5 | 43 | 17 | 71.67 | 28.33 |
| GPT-4 | 44 | 16 | 73.33 | 26.67 |
| LLaMa2 | 40 | 20 | 66.67 | 33.33 |
| Mixtral | 44 | 16 | 73.33 | 26.67 |
| | | Average | 72 | 28 |
| **GPT-4 vs BKG (Second Scenario)** | | | | |
| Claude | 39 | 21 | 65.00 | 35.00 |
| GPT-3.5 | 52 | 8 | **86.67** | 13.33 |
| GPT-4 | 47 | 13 | 78.33 | 21.67 |
| LLaMa2 | 34 | 26 | 56.67 | 43.33 |
| Mixtral | 41 | 19 | 68.33 | 31.67 |
| | | Average | 71 | 29 |
| **Clinician vs BKG (Third Scenario)** | | | | |
| Claude | 48 | 12 | 80.00 | 20.00 |
| GPT-3.5 | 49 | 11 | 81.67 | 18.33 |
| GPT-4 | 55 | 5 | **91.67** | 8.33 |
| LLaMa2 | 44 | 16 | 73.33 | 26.67 |
| Mixtral | 51 | 9 | 85.00 | 15.00 |

|  | | Average | 82.33 | 17.66 |
| --- | --- | --- | --- | --- |
| **Clinician vs BKG-GPT (Fourth Scenario)** | | | | |
| Claude | 50 | 10 | 83.33 | 16.67 |
| GPT-3.5 | 52 | 8 | 86.67 | 13.33 |
| GPT-4 | 56 | 4 | **93.33** | 6.67 |
| LLaMa2 | 48 | 12 | 80.00 | 20.00 |
| Mixtral | 52 | 8 | 86.67 | 13.33 |
| | | Average | **86** | 14 |

**Stage 2: Evaluation of 1500 Differential Diagnoses by Biomedical Knowledge Graph+GPT-4**

Based on the inference from the previous step, we evaluated all 1500 predictions from five LLMs using a combination of Biomedical Knowledge Graphs and GPT-4 (BKG-GPT). Table 4 provides a detailed performance comparison of LLMs in generating differential diagnoses with and without laboratory data. The table provides detailed insight into the performance of five language models (Llama-2, Claude-2, Mixtral, GPT-3.5, and GPT-4) across three scenarios: Top 1, Top 5, and Top 10 differential diagnoses. Metrics include Exact Match, Relevant diagnoses, Incorrect diagnoses, Accuracy, and Lenient Accuracy. Table 2 presents the results for both scenarios with and without lab data.

GPT-4 generally performed the best across multiple scenarios, particularly in terms of lenient accuracy. It achieved the highest lenient accuracy for the Top 1 (74.5%), Top 5 (78.5%), and Top 10 (80%) differential diagnoses with lab data. These results highlight GPT-4's strong ability to generate relevant diagnoses across the differential lists. Mixtral also demonstrated strong performance, particularly in the Top 5 and Top 10 scenarios with lab data. For the Top 5 diagnoses, Mixtral achieved an accuracy of 60% and a lenient accuracy of 80%. In the Top 10 diagnoses, it maintained an accuracy of 58% with a lenient accuracy of 79%, showing consistently high performance across larger diagnosis lists. When comparing the top two performing LLMs, GPT-3.5 showed notable success in generating Top 5 differential diagnoses with lab data, where it achieved the highest lenient accuracy of 77% and a solid accuracy of 54%. This reflects a very low error rate in providing relevant diagnoses within the top 5 predictions, demonstrating its capability in handling key clinical scenarios. GPT-3.5 was also the only model to achieve more than 50% accuracy across all three DDx scenarios. Claude-2 and LLaMa-2 displayed comparable performances overall, though Claude-2 had a slight edge over LLaMa-2 in several metrics. For instance, Claude-2 had a higher exact match rate in the Top 1 differential diagnosis (5 exact matches vs. LLaMa-2's 3) with lab data. Additionally, Claude-2 achieved higher accuracy and lenient accuracy in the Top 5 and Top 10 scenarios. Specifically, Claude-2's Top 5 diagnoses had an accuracy of 58% and a lenient accuracy of 79%, while in the Top 10, it maintained an accuracy of 58% and a lenient accuracy of 79%.

For Top 1 DDx with lab data, GPT-4 had the highest exact match rate, achieving 8 exact matches, while Mixtral followed with 7, and Claude-2 with 5. LLaMa-2 predicted 3 exact matches, and GPT-3.5 lagged with 2. Without lab data, GPT-4 achieved 1 exact match, while Claude-2, Mixtral, and GPT-3.5 did not achieve any exact matches. LLaMa-2 also did not predict any exact matches in this scenario. For Top 5 DDx with lab data, Mixtral performed the best, achieving 10 exact matches, while Claude-2 followed closely with 8. GPT-4 achieved 7 exact matches, while GPT-3.5 and LLaMa-2 both predicted 4 exact matches. Without lab data, GPT-4, Claude-2, and Mixtral achieved 2 exact matches each, while LLaMa-2 and GPT-3.5 had 2 and 1 exact matches, respectively. Finally, for Top 10 DDx with lab data, GPT-4 had the highest exact match rate, achieving 10 exact matches. Claude-2 and Mixtral both followed with 8 exact matches, while LLaMa-2 and GPT-3.5 achieved 4 and 3 exact matches, respectively. Without lab data, Claude-2 led with 4 exact matches, GPT-4 followed with 2, and Mixtral and GPT-3.5 both had 2 and 3 exact matches, respectively. LLaMa-2 predicted 3 exact matches in this scenario. The ability to achieve a high exact match rate in differential diagnosis is crucial, as it directly reflects the model's capacity to provide the most accurate and relevant diagnosis, significantly enhancing clinical decision-making efficiency and reliability.

In summary, GPT-4 consistently showed the highest performance in generating differential diagnoses across various scenarios, particularly excelling in lenient accuracy. Mixtral followed closely, especially in the Top 5 and Top 10 lists, while GPT-3.5 stood out for its high lenient accuracy in the Top 5 diagnoses. Claude-2 slightly outperformed LLaMa-2, particularly in exact match rates and lenient accuracy across different differential diagnosis lists.

**Table 4.** The accuracy of five LLMs in generating differential diagnosis lists of Top 1, Top 5, and Top 10 from 50 case reports automatically evaluated by BKG-GPT. Accuracies and lenient accuracy were calculated using equations 1 and 2 respectively.

|  | Top 1 Differential Diagnosis (with lab) | | | | | Top 1 Differential Diagnosis (without lab) | | | | |
|---|---|---|---|---|---|---|---|---|---|---|
| LLM | Exact Match | Relevant | Incorrect | Accuracy | Lenient accuracy | Exact Match | Relevant | Incorrect | Accuracy | Lenient accuracy |
| Llama-2 | 3 | 46 | 1 | 52% | 75% | 0 | 49 | 1 | 49% | 73.5% |
| Claude-2 | 5 | 40 | 5 | 50% | 70% | 0 | 43 | 7 | 43% | 64.5% |
| Mixtral | 7 | 38 | 5 | 52% | 71% | 0 | 43 | 7 | 43% | 64.5% |
| GPT-3.5 | 2 | 40 | 8 | 44% | 64% | 0 | 42 | 8 | 42% | 63% |
| GPT-4 | 8 | 39 | 3 | 55% | 74.5% | 1 | 41 | 8 | 43% | 63.5% |
|  | Top 5 Differential Diagnosis (with lab) | | | | | Top 5 Differential Diagnosis (without lab) | | | | |
| Llama-2 | 4 | 46 | 0 | 54% | 77% | 2 | 47 | 1 | 51% | 74.5% |
| Claude-2 | 8 | 42 | 0 | 58% | 79% | 1 | 47 | 2 | 49% | 72.5% |
| Mixtral | 10 | 40 | 0 | 60% | 80% | 1 | 47 | 2 | 49% | 72.5% |
| GPT-3.5 | 4 | 46 | 0 | 54% | 77% | 2 | 41 | 1 | 51% | 74.5% |
| GPT-4 | 7 | 43 | 0 | 57% | 78.5% | 2 | 47 | 1 | 51% | 74.5% |
|  | Top 10 Differential Diagnosis (with lab) | | | | | Top 10 Differential Diagnosis (without lab) | | | | |
| Llama-2 | 4 | 46 | 0 | 54% | 77% | 3 | 46 | 1 | 52% | 75% |
| Claude-2 | 8 | 42 | 0 | 58% | 79% | 4 | 45 | 1 | 53% | 75.5% |
| Mixtral | 8 | 42 | 0 | 58% | 79% | 2 | 47 | 1 | 51% | 74.5% |
| GPT-3.5 | 3 | 47 | 0 | 53% | 76.5% | 3 | 45 | 2 | 51% | 73.5% |
| GPT-4 | 10 | 40 | 0 | 60% | 80% | 2 | 48 | 0 | 52% | 76% |

Fig. 3 shows the effect of including lab test results in clinical vignettes across different models, illustrating that the inclusion of lab results data enhances both accuracy and lenient accuracy for all models. GPT-4 consistently achieved the highest accuracy and lenient accuracy, particularly when lab data was included, demonstrating its superior ability to integrate clinical information. Mixtral and Claude-2 performed strongly, showing significant improvements in both accuracy and lenient accuracy when lab results were included. Mixtral achieved the highest performance in Top 5 DDx and after GPT-4 in the Top 1 and Top 10, followed closely by Claude-2, particularly in the Top 5 and Top 10 categories. In contrast, GPT-3.5, while showing improvements with the inclusion of lab data, did not perform as well as Mixtral and Claude-2. LLaMa-2 achieved the highest accuracy in the Top 1 and higher accuracy than GPT-3.5 in the Top 5 and Top 10 categories. These results emphasize the critical role of lab data in improving diagnostic accuracy, with GPT-4 leading the models, followed by Mixtral and Claude-2.

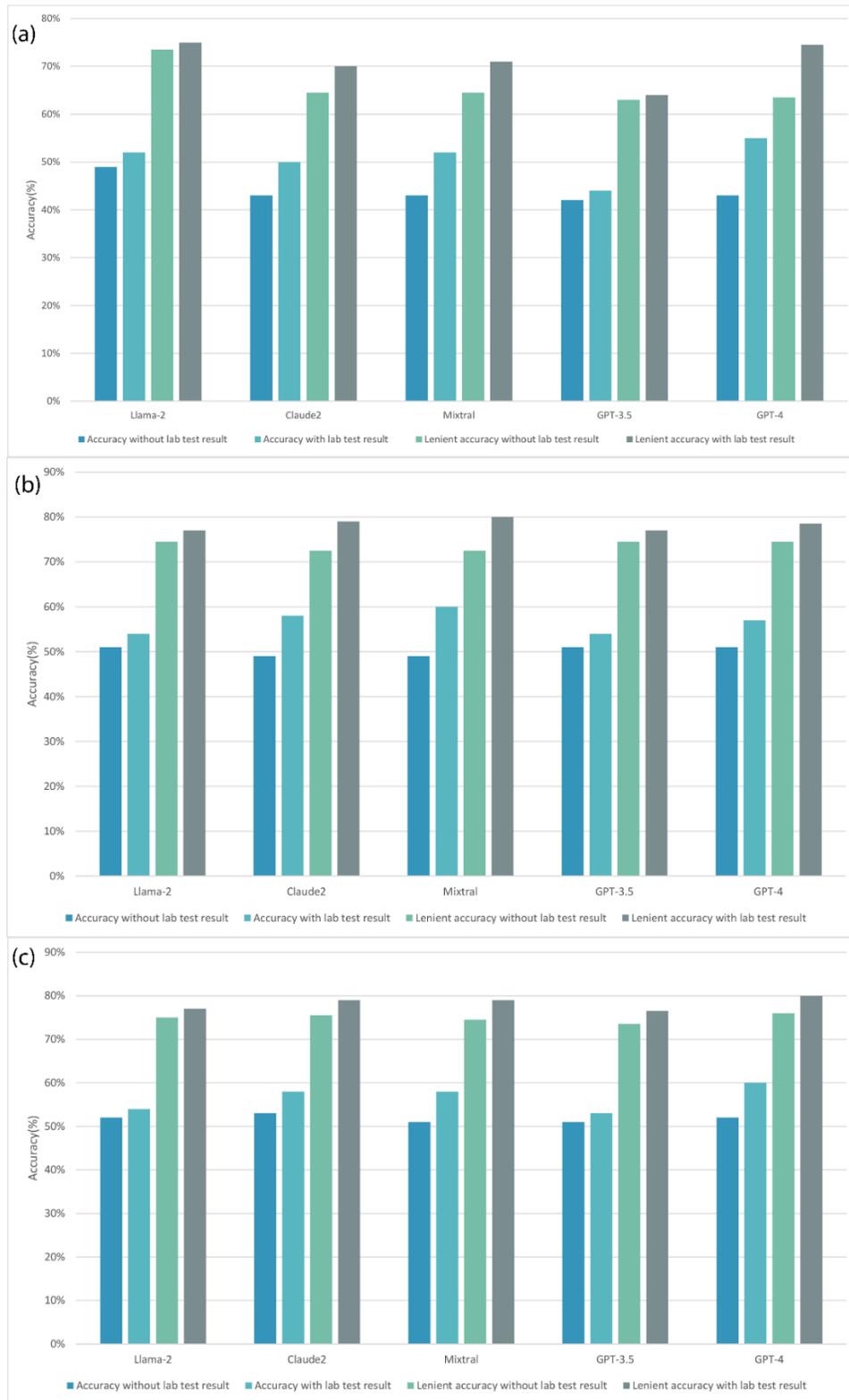

**Fig. 3. Accuracy and lenient accuracy of LLMs including and excluding lab test results for (a) Top differential diagnosis, (b) Top 5 differential diagnosis, and (c) Top 10 differential diagnosis.**

Table 5 compares p-values from paired t-tests to evaluate the impact of incorporating lab test data on model performance across three prediction categories—Top 1, Top 5, and Top 10—using two metrics: accuracy and lenient accuracy. All p-values are below 0.05, indicating statistically significant differences between models with and without lab data. For the accuracy metric, the p-values (e.g., 0.023 for Top 1) suggest that lab data meaningfully enhances the precision of the models. Similarly, in the more flexible lenient accuracy metric, significant improvements are observed, with the Top 10 predictions showing the strongest effect (p = 0.001). The results highlight that lab test data can significantly impact the models' accuracy.

**Table 5: Comparison of P-values from Paired T-tests evaluating the impact of Lab test data**

| With and Without lab test | P Value (Accuracy) | P Value (Lenient accuracy) |
|---|---|---|
| Top 1 | 0.023 | 0.049 |
| Top 5 | 0.016 | 0.011 |
| Top 10 | 0.018 | 0.001 |

*P values are from paired t-tests.

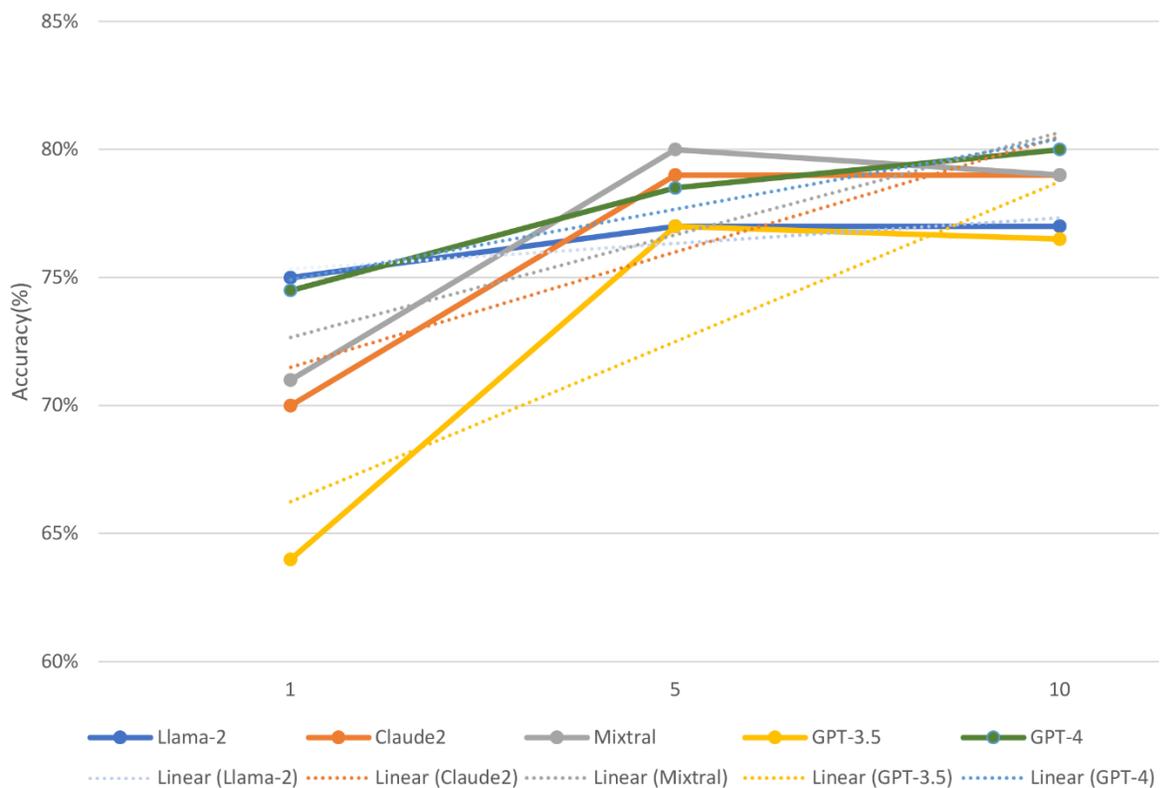

**Fig. 4. Accuracy and linear accuracy of the LLMs for DDx with lab test data.**

Fig. 4 presents the accuracy and linear trend lines of five LLMs—LLaMa-2, Claude-2, Mixtral, GPT-3.5, and GPT-4—across the Top 1, Top 5, and Top 10 differential diagnoses with lab test results. The graph shows that GPT-4 consistently maintains the highest accuracy across all three categories, achieving 55% accuracy for the Top 1 diagnosis, 57% for the Top 5, and 60% for the Top 10 when lab results are included, with lenient accuracies ranging from 74.5% to 80%. The linear trend lines reflect overall improvement in accuracy as the number of diagnoses increases, with GPT-4, Mixtral, and Claude-2 showing the strongest upward trends, especially when lab data is included. In contrast, LLaMa-2's trend remains relatively flat, highlighting its challenges in achieving higher accuracy even with additional clinical information.

Mixtral and Claude-2 also performed well, particularly in the Top 5 and Top 10 categories with lab results. Mixtral achieved 60% accuracy in the Top 5 diagnoses and 58% in the Top 10, with lenient accuracies of 80% and 79%, respectively. LLaMa-2 achieved the highest lenient accuracy for the Top 1 diagnosis with lab data at 75%, which highlights its ability to provide relevant diagnoses in this category. However, in terms of overall accuracy, LLaMa-2 reached 52% in the Top 1 diagnosis with lab data and showed moderate improvements with increasing diagnosis lists, achieving 54% accuracy in the Top 10 diagnoses. Despite showing some improvements as more diagnoses were considered, it remained one of the lower-performing models overall in terms of accuracy, though its lenient accuracy results were more competitive, especially in specific categories like Top 1 diagnoses with lab data.

**Error Analysis**

The LLM models showed varying error rates (incorrect diagnoses). For Top 1 differential diagnosis with lab data, Llama-2 had the lowest error rate (1 incorrect diagnosis), while GPT-3.5 had the highest (8 incorrect diagnoses). The error rates were zero for Top 5 and 10 DDx among all LLMs with lab data showing the predictions have some meaningful connection with the final diagnosis. Without lab data, the error rates increased across all models. For instance, in the Top 1 scenario without lab data, GPT-4 and GPT-3.5 had 8 incorrect diagnoses, while Mixtral had 7 incorrect diagnoses. Exact match rates were relatively low across all models, indicating the difficulty of achieving an exact diagnosis match. For example, in Top 1 differential diagnosis with lab data, GPT-4 had the highest exact match rate (8), whereas others had lower exact match rates. Lenient accuracy rates were significantly higher than exact match rates, reflecting the models' ability to provide relevant but not exact diagnoses. This indicates that the models are better at providing useful differential diagnoses rather than pinpointing the exact one every time.

The sub-set of 300 differential diagnosis comparisons between the predictions made by LLMs and their evaluations by GPT-4, clinicians, and the BKG reveals nuanced insights into the accuracy and alignment of these models with clinical standards. The evaluation of LLM GPT-4 reveals that its predictions predominantly align with both clinician comments and BKG evaluations as "Relevant" or "Exact Match," demonstrating its reliability in producing clinically accurate information. In the case report (PMID 19162360, final diagnosis: *diabetic nephropathy with near-nephrotic range proteinuria*), GPT-4's prediction (diabetic nephropathy) was "Relevant," and this assessment was supported by both clinician comments and BKG, indicating close alignment with clinical standards. For LLM GPT-3.5, a similar pattern of alignment is observed. Most of its "Relevant" predictions were consistently categorized as such by both GPT-4 and clinicians, as seen in the example of PMID 23415437 (Final diagnosis: *Coronary heart disease (CHD) caused by ApoA-INashua mutation* vs GPT-3.5 diagnosis: *Hypoalphalipoproteinemia (low HDL-C) secondary to a novel heterozygous A-1 in-frame insertion mutation*). This indicates that GPT-3.5 frequently provides relevant diagnoses that align well with clinical evaluations. However, occasional discrepancies arise, highlighting the importance of integrating these LLM predictions with systematic data sources like BKG to enhance diagnostic accuracy.

For LLM Claude, the analysis shows a high degree of consistency between GPT-4's evaluations and clinician comments, particularly in scenarios where predictions were categorized as "Exact Match." For instance, in the case reports with PMID 31497118 (Final diagnosis: *Drug-induced liver injury (DILI) from levetiracetam (LEV)* vs Claude diagnosis: *Drug-induced liver injury caused by levetiracetam(LEV)*) and PMID 31497445 (Final diagnosis: *Vancomycin-induced DRESS syndrome* vs Claude diagnosis: *Vancomycin-induced drug reaction with eosinophilia and systemic symptoms(DRESS) syndrome*), Claude's predictions were marked as "Exact Match" by both GPT-4 and clinicians, indicating strong alignment . This suggests that Claude's predictions align well with clinical judgments when they are accurate. Additionally, Claude LLM generated incorrect predictions, in the case of PMID 31380008 (Final diagnosis: *AMAN subtype of Guillain-*

*Barré syndrome* vs Claude: *Acute Hepatitis A Infection*), Claude's prediction was deemed "Incorrect" by both GPT-4 and clinicians, further demonstrating consistency in identifying less accurate predictions. When comparing the evaluations across GPT-4, clinicians, and BKG for LLM Claude, most predictions categorized as "Relevant" were similarly evaluated as "Relevant" by all three evaluators, reinforcing the model's ability to produce clinically meaningful information. However, there are instances where "Exact Match" outcomes were achieved, demonstrating that while Claude often aligns with clinician judgment, exact matches are rare but significant when they occur. The evaluations of LLaMA2 and Mixtral present more variability. For example, while both models show strong alignment with "Relevant" predictions, they tend to diverge more frequently when aiming for "Exact Match" outcomes, suggesting less consistency in achieving the highest levels of diagnostic accuracy compared to Claude and GPT-3.5. Overall, the analysis underscores the importance of integrating LLM-generated predictions with systematic data sources like BKG and clinician evaluations to enhance diagnostic accuracy. While exact matches are rare across all models, the consistent identification of "Relevant" predictions highlights the utility of these models in providing clinically valuable insights that align well with clinician judgment and data-driven assessments.

For a subset of LLM predictions, we asked LLM to generate the reasoning behind each prediction and asked clinicians to evaluate it. Clinicians assessed whether the provided reasoning was logical and whether it supported the accuracy of the diagnosis. For the "Guillain-Barr Syndrome following acute hepatitis E diagnosis", it correctly predicted, "Acute hepatitis E complicated by hepatic encephalopathy and Guillain-Barr syndrome (GBS)". The model provided the reason for the prediction as: "*Initial and subsequent lab tests showed elevated liver enzymes ( alanine aminotransferase (ALT) and aspartate aminotransferase (AST)), and very high bilirubin levels, which are indicative of acute liver injury. The serology tests confirmed the presence of HEV antibodies (both IgM and IgG). The prolonged presence of IgM anti-HEV indicates a sustained immune response to the infection. The development of lower limb weakness progressing to severe muscle weakness represents a rare but recognized neurological manifestation associated with HEV infection. The rapid progression and the timing of these symptoms during the course of the liver disease further link them to the HEV infection.*" The reasoning was evaluated by the clinician as "solid decision making". Another example is for the final diagnosis "Aspergillus fumigatus infection with liver failure", and the model predicted "Acute liver failure complicated by invasive pulmonary aspergillosis". The model reasoning was that '*the thoracic CT scan showing bilateral pulmonary high-density shadows suggests a significant pulmonary involvement, which is not typical of liver diseases alone and the markedly elevated bilirubin levels and liver enzymes (alanine aminotransferase and aspartate aminotransferase) indicate acute liver injury. The progression to a high INR (International Normalized Ratio) and worsening bilirubin levels despite treatment suggest acute liver failure*'. The reasoning was also deemed correct by the clinician.

GPT-4 model also made some incorrect predictions, for example, for the final diagnosis of "Ondansetron (OSDN)-induced hepatotoxicity", the model predicted "Gestational Hyperemesis compounded by a Drug-induced Hepatotoxicity and a Pregnancy-related Psychotic Disorder". The reasoning provided was, *"The patient's symptoms of persistent nausea, vomiting, and significant weight loss over four weeks are indicative of hyperemesis gravidarum, especially given their occurrence during pregnancy. The improvement in these symptoms post-delivery further supports this diagnosis, as hyperemesis gravidarum typically resolves after delivery. Drug-induced Hepatotoxicity: The marked elevation in liver enzymes (AST and ALT), which escalated during treatment and began to normalize post-delivery, suggests a drug-induced liver injury. Olanzapine, initially used, is known for potential hepatotoxic effects, which could explain the initial liver enzyme elevations"*. This was a wrong prediction and the clinician confirmed it with a comment "the LLM missed the resolution of symptoms and labs with the stopping of the medicine". For a correct prediction of "Thallium intoxication", for a case report diagnosis "Acute thallium poisoning" the Mixtral model provided a reason *"The patient's plasma and 24-hour urine tests for heavy metals show a significantly elevated level of thallium in the urine (540 µg/g Cr) compared to the normal range (0.4-10 µg/g Cr) and a normal level of thallium in the plasma (12 µg/dL) which is consistent with the diagnosis of thallium poisoning. Thallium is a highly toxic heavy metal that can cause a wide range of symptoms, including gastrointestinal disturbances, neurologic symptoms, and liver damage"*. The clinician commented that, "This is sound decision making, arriving at the right diagnosis. The details of labs made this clear".

Generally, the models were able to interpret the lab test results along with symptoms and medications administered. The models generated solid interpretations like, 'The lab test results show elevated total bilirubin, direct bilirubin, aspartate transaminase, alanine transaminase, prothrombin time, international normalized ratio, lactate dehydrogenase, and C-reactive protein, which can be indicative of liver damage and inflammation, which can be seen in heavy metal poisoning.' and the clinician commented as 'This is sound decision making, arriving at the right diagnosis. The details of the labs made this clear.

**Case incidence**

We conducted a search for the incidence of diseases corresponding to the final diagnosis in the 50 cases by reviewing the literature on PubMed (https://pubmed.ncbi.nlm.nih.gov/). The search results provided in Table 6 revealed a wide range of disease incidence, with the majority of cases being reported in fewer than 100 articles. Specifically, 22 diagnoses had 1-10 articles, and 13 diagnoses had 11-100 articles, highlighting the rarity of these conditions. Additionally, there were 10 diagnoses with 101-1000 articles, indicating these diseases are relatively uncommon. The search results show that 70% of the diagnoses (35 out of 50) have fewer than 100 articles in PubMed, emphasizing their rarity. In contrast, 20% of the diagnoses (10 out of 50) fall within the 101 to 1000 report range, suggesting these are somewhat more prevalent but still uncommon. Meanwhile, 8% of the diagnoses (4 out of 50) have more than 10000 articles, showing these are more frequently occurring or well-documented conditions in the literature. This highlights the rarity of the majority of the diagnoses, as 70% of them are reported in fewer than 100 articles. Since these are such rare conditions, LLMs must possess specific knowledge of these diseases to make accurate diagnosis predictions.

**Table 6**. Disease Incidence Distribution Based on PubMed Literature Review for 50 Case Reports

| Disease incidence range | Number of cases |
| --- | --- |
| 1-10 | 22 |
| 11-100 | 13 |
| 101-1000 | 10 |
| 1001-10000 | 1 |
| >10000 | 4 |

# Discussion

This study evaluated the impact of lab test results on the accuracy of differential diagnoses using five LLMs with published clinical case reports. The results showed that including lab data improved both accuracy and lenient accuracy, with GPT-4 achieving the highest performance in generating relevant differential diagnoses, even if the exact match was not always achieved. The

Mixtral-8x7B model also performed well, particularly with lab data, highlighting the advanced capabilities of these LLMs in processing complex clinical information. A detailed analysis of the 300 selected predictions evaluated using GPT-4, BKG and clinicians, along with their various combinations, revealed that GPT-4's predictions align significantly better with clinicians when relevant predictions are considered as correct. This improvement underscores the practical utility of GPT-4's predictions, even if they are not exact matches. Furthermore, the combination of GPT-4 and BKG evaluations achieved the highest accuracy, indicating that integrating LLM-generated predictions with systematic data enhances the relevance and clinical utility of diagnostic predictions.

The high performance of GPT-4 (lenient accuracy: 74% - 80%) indicates a strong ability to provide relevant differential diagnoses even if the exact match is not achieved. The consistent performance (lenient accuracy: 71% - 80%) of Mixtral suggests that it is reliable in providing a broader set of relevant differential diagnoses. It is also worth mentioning that GPT-4 is the highest-performing LLM in predicting the exact diagnosis of 8 and 10 cases in the Top 1, and 10 DDx list and Mixtral predicted 10 exact cases in the Top 5 DDx respectively. GPT-4 excelled and achieved the best performance by predicting most of the relevant DDx list. Accuracy and lenient accuracy were generally higher when lab data was included, highlighting the importance of lab data in improving diagnostic accuracy. Incorporating lab data significantly enhances model performance, with GPT-4's accuracy increasing from 43% to 55% in the Top 1 scenario, representing a 12% improvement and 7 additional exact case diagnoses. GPT-4 stands out for its balanced performance across all scenarios, suggesting its robustness and reliability in clinical decision support. Mixtral's consistent performance in providing relevant diagnoses makes it a reliable option for scenarios where exact matches are less critical. Claude-2 and Llama-2, while slightly behind GPT-4 and Mixtral, still show competent performance, particularly when lab data is available.

Clinicians generally observed that model performance varied based on case complexity and lab requirements. One clinician commented that in the simpler case, with a final diagnosis of "diabetic nephropathy with near-nephrotic range proteinuria" (PMID: 19162360) most models missed the near-nephrotic proteinuria, and the relevance of differentials decreased as the number of diagnoses expanded (DDx5 and DDx10). In the moderately complex case, with a final diagnosis of "coronary heart disease (CHD) caused by ApoA-I Nashua mutation" (PMID: 23415437) models often failed to connect the genetic mutation causing the lipid disorder to coronary artery disease. In the most complex case, diagnosed as "Stage IV classical Hodgkin's lymphoma" (PMID: 23975921) involving multiple specialists and extensive diagnostic procedures, model predictions were less accurate. Overall, as case complexity increased and more specialized labs were needed, model predictions became less precise and struggled to link related diagnoses comprehensively. Additionally, models often redundantly included diagnoses already confirmed in the case study, a limitation that could be addressed with more advanced techniques in LLMs. One clinician also pointed out that the Top 5 and Top 10 DDx made by LLMs got more irrelevant compared by what a human clinician would make.

The superior performance of the GPT-4 and Mixtral-8x7B model, across different scenarios, underscores the advanced capabilities of the latest LLMs in processing and integrating complex clinical data for diagnosis. However, the observed performance dropped when lab test results were excluded and for complex diseases with lab tests raises important considerations for implementing LLMs in clinical settings. The slight lag in the performance of Mixtral-8x7B compared to GPT-4, for instance, offers a starting point for further research such as Retrieval Augmented Generation (RAG) for medical applications.

## Conclusions

Through the evaluation of five LLMs (GPT-4, GPT-3.5, Llama-2, Claude2, and Mixtral-8x7B) on the clinical case reports from PMC-Patients dataset, the study reports that the accuracy of differential diagnoses improves substantially when lab test results are included, underscoring their critical role in accurate medical diagnosis. The inclusion of lab test results significantly enhances the accuracy and lenient accuracy of differential diagnosis predictions made by large language models, especially in improving the exact match predictions.  Lab data, such as liver function tests, toxicology/metabolic panels, and serology/immune tests, were generally interpreted correctly, enhancing the models' ability to generate relevant diagnoses. The study also found that the combination of Biomedical Knowledge Graphs and GPT-4 (BKG-GPT) can perform automatic assessments with a level of accuracy comparable to that of clinicians. Our study demonstrates that models such as GPT-4 and Mixtral-8x7B excel in providing relevant differential diagnoses when lab data is considered, with GPT-4 achieving the highest lenient accuracy across various scenarios. Although exact match rates remain relatively low, the high performance in lenient accuracy suggests that these models are adept at generating plausible diagnoses, thus offering valuable support in clinical decision-making. The findings underscore the critical role of lab data in improving diagnostic precision and the advanced capabilities of current LLMs in integrating complex clinical information.

## Methods

### Study Design

We assessed the impact of laboratory test results on enhancing the accuracy of differential diagnosis using five large language models: GPT-4[14], GPT-3.5[23], Llama2-70B[24], Claude-2[25], and Mixtral 8x7B-Chat[26]. Clinical case reports were sourced from the PMC-Patients dataset for this evaluation. The term "differential diagnosis" refers to a list of potential conditions or diseases that may be causing a patient's symptoms and signs. Clinicians consider the patient's clinical history, physical examination findings, and investigation results, collectively known as clinical vignettes, to aid in the diagnostic process. For this study, clinical vignettes were manually generated from 50 selected case reports, including details such as the patient's age, gender, symptoms, laboratory test results, and other relevant information, to enable the language models to formulate differential diagnostic responses. Initially, clinical reports including laboratory test results and age-specific cases, were selected to generate clinical vignettes. These vignettes were manually extracted from the clinical case presentation sections, encompassing details such as age, sex, symptoms, full case report, and lab tests. A differential diagnosis (DDX) prompt was then created, instructing the large language models to consider all pertinent details and formulate comprehensive and accurate differential diagnoses. These predictions were categorized into top 10, top 5, and top 1 differential diagnoses. The accuracy of the model's predictions was evaluated using metrics such as exact match, relevance, and incorrect predictions, with accuracy further divided into exact and lenient categories.

### Ethical Considerations

Since this study employed case vignettes derived from publicly available published case reports, approval from the ethics committee and the requirement for individual consent were not necessary.

### Clinical Vignettes

We utilized PMC-Patients[22], a novel benchmark dataset that includes patient summaries and relationships derived from PubMed Central articles, to collect 50 clinical case reports. PMC-Patients encompasses 167,000 patient summaries with 3.1 million patient-article relevance annotations and 293,000 patient-patient similarity annotations, making it the largest resource for ReCDS and one of the largest patient collections available. Case reports were manually selected to cover a wide range of diseases such as Endocrine/Metabolic, Cardiovascular, Hematologic/Oncologic, Infectious Diseases Neurological Disorders, etc., ensuring equal representation of genders and various age groups. Figure 2 illustrates the distribution of diseases across various clinical categories. Following the selection of 50 case reports, four undergraduate premedical students were recruited to manually extract details such as age, sex, symptoms, lab tests, full case report, and final diagnosis to generate the clinical vignettes.

For example, consider the case report titled "Acute cytomegalovirus hepatitis in an immunocompetent host" (PMID: 24275336)[27]. From this case report we extracted the following data, *Age*: '52', *Gender*: 'Female', *Lab test*: 'serum aspartate aminotransferase of 739 U/L (normal value 15-37 U/L)....', *Case Report*: 'A 52-year-old Hispanic woman with a medical history of hypoparathyroidism....' *Final diagnosis*: 'Acute cytomegalovirus hepatitis'. All the case reports are indexed in PubMed and published in peer-reviewed clinical journals. The final diagnosis for each case was established through standard diagnostic processes and subsequently documented in these case reports.

**Differential Diagnosis Lists generated by LLMs**

We utilized several large language models for our study: GPT-4 and GPT-3.5 (OpenAI, LLC), Llama-2-70b-chat (Meta LLC), Claude 2 (Anthropic, LLC), and Mixtral 8x7B Mixture-of-Experts (Mistral AI, LLC). None of these models were specifically trained or reinforced for medical diagnoses. We accessed the GPT models through the OpenAI GUI (https://chatgpt.com/), while the Llama-2, Claude-2, and Mixtral 8x7B models were accessed via the open-source web interface POE (https://poe.com/). To ensure no influence from previous interactions, each model was prompted with a fresh chat interface. The initial prompt used was: "Imagine you are a Medical Professional tasked with providing one (1) comprehensive and accurate diagnosis for a patient presenting with the following case report. Please consider the patient's Age, Gender, Symptoms, Lab tests, and the full Case Report and any pertinent details to formulate your response." This prompt was followed by the clinical vignette as described earlier. To generate five and ten differential diagnoses (DDx), the prompts were adjusted to request "five (5) comprehensive and accurate differential diagnoses" and "ten (10) comprehensive and accurate differential diagnoses," respectively. For evaluating the role of lab tests, DDx were generated both including and excluding the laboratory test results, starting with the prompt excluding lab test results. The final prompt was refined using prompt engineering techniques and by evaluating various prompts to encourage the LLMs to generate comprehensive lists of DDx. This optimized prompt template consistently yielded reliable and inclusive differential diagnoses across all the LLMs.

**Evaluation of Differential Diagnosis Lists**

The current study design generates 1,500 differential diagnosis (DDx) sets, comprising 50 case reports evaluated by five large language models (LLMs) across six conditions. The six different conditions are Top 1, Top 5, and Top 10, with each considered both with and without lab test results. To comparatively evaluate the LLMs' ability to generate a DDx, we designed a two-stage evaluation process as follows:

Stage 1: Evaluation by medically trained clinicians, Biomedical Knowledge Graph (BKG), and GPT-4. We selected 300 predictions (20% of our total 1,500 predictions) from 10 case reports and asked clinicians to assess the LLM-generated diagnoses against the actual diagnoses. This was compared with GPT-4 and a BKG.

Stage 2: Fully automated evaluation of all the 1500 predictions from 5 LLMs using GPT-4 combined with 2-hop subgraph extraction and PageRank-based similarity computation on a BKG.

**GPT-4 based evaluation**: For evaluation, we post-processed the LLM outputs, generated a JSON file, and used API calls to GPT4 for LLM-based evaluation. The full evaluation pipeline including the prompt is provided in Fig. 5. The evaluation metrics are defined as follows:

Exact Match: The predicted diagnosis is the same as the true diagnosis.

Relevant: The predicted diagnosis is a variant, form, or closely related term referring to the same condition. It captures the broad category or concept of the true diagnosis but may differ in specifics.

Incorrect: The predicted diagnosis does not accurately reflect the true diagnosis.

A score is assigned to the metrics namely 'Exact Match:1.0', 'Relevant:0.5', and 'Incorrect:0.0' by comparing the actual and predicted diagnosis.

For example, in Figure 5, the predicted diagnosis by an LLM is 'diabetic nephropathy' and the actual diagnosis from the case report is 'diabetic nephropathy with near nephrotic range proteinuria'. The actual diagnosis is a sub-type of the predicted diagnosis, not an exact match but still a relevant prediction. GPT-4 predicted the relevant of this example.

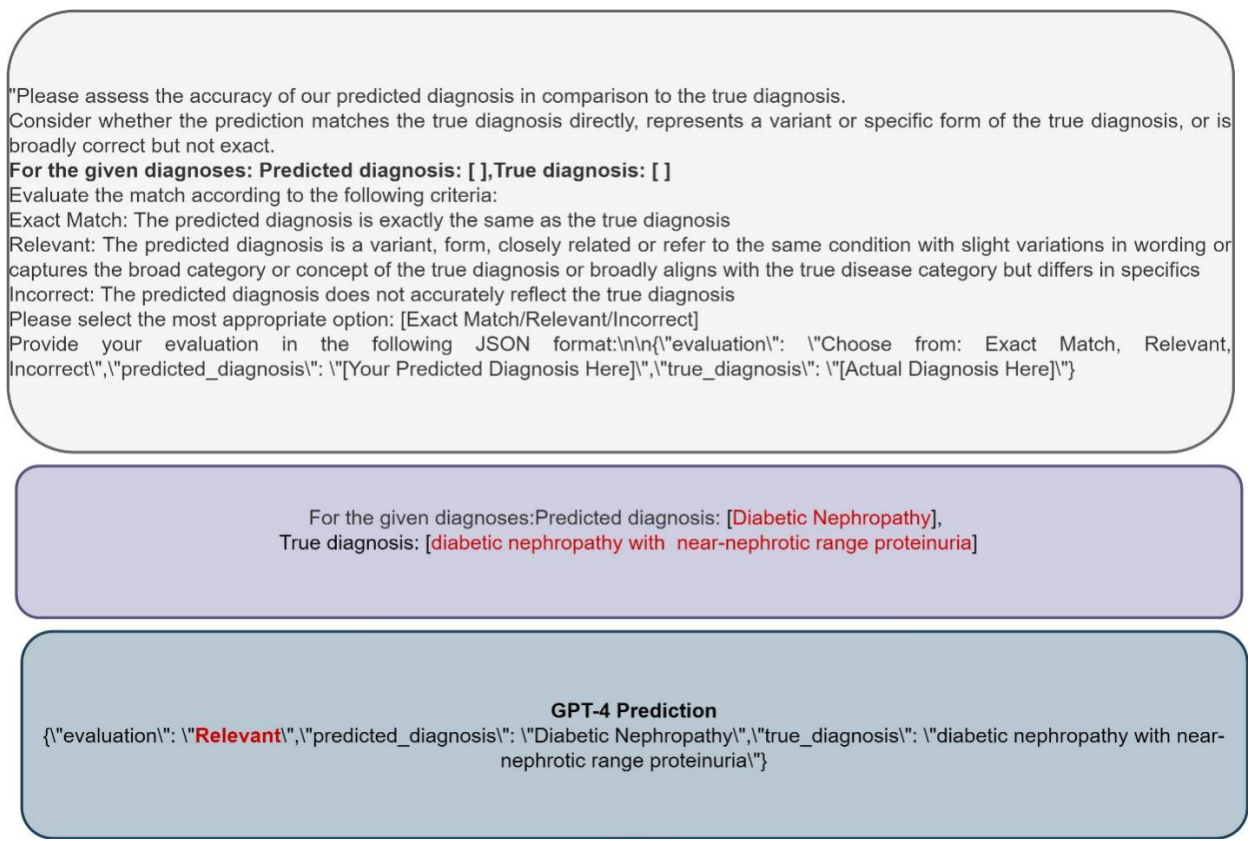

**Fig. 5. Example of automatic evaluation differential diagnosis from LLMs using GPT-4**

**Knowledge Graph-based evaluation**

In this study, we utilized the PromptLink method[28] to link diagnosis entities into external BKG for evaluation. PromptLink is a novel and highly accurate framework designed for biomedical concept linking across diverse data sources without requiring prior knowledge, context, or training data.

Specifically, PromptLink employs a biomedical-specialized, pre-trained language model (SapBERT[29]) to generate BKG concept candidates for each diagnosis entity. Subsequently, a large language model (GPT-4) is used to establish linkages between diagnosis entities and BKG concepts through two-stage prompts. This process enables the linkage of diagnosis entities to the closest concept in the external BKG, facilitating BKG-based evaluation.

The evaluation was conducted using the iBKH BKG, which comprises 2,384,501 biomedical entities[30]. To reduce the computational cost, we extracted BKG concepts with more than five neighboring nodes from iBKH, specifically focusing on entities related to drugs, diseases, anatomy, side effects, symptoms, and therapeutic classes, while excluding other concepts. For each true diagnosis entity, we identified the linked BKG concept node as the center node and sampled a subgraph that included all two-hop neighbors and their relations from the BKG. Within this subgraph, the PageRank value for each node was computed by using the tool "igraph"[31]. The predicted diagnosis entities were also linked to the corresponding BKG concepts, and the matching score for each pair of true diagnosis and predicted diagnosis was calculated as follows:

Matching score of 3: The predicted diagnosis entity exactly matches the true diagnosis entity.

Matching score of 2: The predicted diagnosis entity is related to the true diagnosis entity, and they are linked to the same node in the BKG.

Matching score of 1: The PageRank value of the predicted diagnosis entity is greater than zero, indicating relevance to the true diagnosis entity within a two-hop distance in the BKG.

Matching score of 0: The PageRank value of the predicted diagnosis entity, which means the predicted diagnosis entity is not relevant to the true diagnosis entity.

We then categorized the BKG-based matching score as follows: 0 for Incorrect, 1 and 2 for Relevant, and 3 for an Exact match.

Based on the evaluation metrics accuracy and lenient accuracy as described in equations 1 and 2 were calculated as the final metrics. Accuracy is calculated by assigning different weights to the types of matches: exact matches are weighted by 1.0, relevant matches by 0.5, and incorrect matches by 0.0. The sum of these weighted values is then divided by the total number of diagnoses evaluated, which is 50. Lenient accuracy is calculated by taking exact matches weighted by 1.0, relevant matches by 0.75, and incorrect matches by 0.0. This sum is then divided by the total number of diagnoses evaluated, which is 50.

$$\text{Accuracy} = \frac{(Exact\ Match * 1.0 + Relevant * 0.5 + Incorrect * 0.0)}{50} \quad\quad (1)$$

$$\text{Lenient Accuracy} = \frac{(Exact\ Match * 1.0 + Relevant * 0.75 + Incorrect * 0.0)}{50} \quad\quad (2)$$

**Acknowledgments**

This work was supported by the AHRQ grant R21HS029969 (PI: ZH). This project was also partially supported by the University of Florida Clinical and Translational Science Institute, which is supported in part by the National Institutes of Health (NIH) National Center for Advancing Translational Sciences under award UL1TR001427. QJ and ZL were supported by the NIH Intramural Research Program, National Library of Medicine.

**Code availability**

The code, results of inferences, and scripts to generate results for all aspects of this study are made publicly available at https://github.com/balubhasuran/DDx_CaseReports

In our experiments, we used Python 3.11, and the following open-source libraries: pandas = 2.2.2, numpy = 1.26.4, and matplotlib==3.9.2

**Author contributions**

Z.H. conceptualized the work. B.B., Q.J., Y.X., C.Y., Z.L., and Z.H. designed the study. B.B., Q.J., Y.X., C.Y., and Z.H. performed the coding and the experiments. B.B. and Z.H. conducted the literature search. Z.L. and Z.H. provided advisory support for the project. K.H., J.C., and C.H. performed the clinical evaluation of the LLM predictions. B.B. and Z.H. prepared the initial draft of the manuscript, with all authors actively participating in the refinement and finalization of the manuscript through comprehensive review and contributions. Z.H. supervised the project.

**Competing interests**

The authors declare no competing interests.